\documentclass{article} %
\usepackage{iclr2024_conference,times}

\definecolor{citecolor}{HTML}{2779af}
\definecolor{linkcolor}{HTML}{c0392b}
\usepackage[pagebackref=false,breaklinks=true,colorlinks,bookmarks=false,citecolor=citecolor,linkcolor=linkcolor]{hyperref}
\usepackage{url}
\usepackage{graphicx}
\usepackage{listings}
\usepackage{inconsolata}

\usepackage{titlesec}
\titlespacing*{\paragraph}{\parindent}{0.25ex}{1ex}
\titlespacing*{\section}{0pt}{3pt}{3pt}
\titlespacing*{\subsection}{0pt}{3pt}{3pt}

\usepackage{amsmath,amsfonts,bm}

\def\Figref#1{Figure~\ref{#1}}

\def\Secref#1{Section~\ref{#1}}

\def\eqref#1{equation~\ref{#1}}

\def\1{\bm{1}}

\DeclareMathAlphabet{\mathsfit}{\encodingdefault}{\sfdefault}{m}{sl}
\SetMathAlphabet{\mathsfit}{bold}{\encodingdefault}{\sfdefault}{bx}{n}

\usepackage{url}
\usepackage{amsmath,amsfonts,amssymb}
\usepackage{graphicx}

\title{A Reply to \cite{aleks2023subspace}'s \\ ``Interpretability Illusion'' Arguments}

\author{Zhengxuan Wu$^{\ast}$ Atticus Geiger$^{\diamondsuit}$ Jing Huang$^{\ast}$ Aryaman Arora$^{\ast}$ \\ \textbf{Thomas Icard$^{\ast}$ Christopher Potts$^{\ast}$ Noah D.~Goodman$^{\ast}$} \\
$^{\ast}$Stanford University $^{\diamondsuit}$Pr(Ai)$^2$R Group \\
\texttt{\{wuzhengx, atticusg, hij, aryamana, icard, cgpotts, ngoodman\}@stanford.edu} \\
}

\newcommand{\Appref}[1]{Appendix~\ref{#1}}

\newcommand{\Eqnref}[1]{Eqn.~\ref{#1}}

\iclrfinalcopy %
\begin{document}

\maketitle

\begin{abstract}
We respond to the recent paper by \cite{aleks2023subspace}, which reviews subspace interchange intervention methods like distributed alignment search (DAS; \citealt{geiger2023finding}) and claims that these methods potentially cause ``interpretability illusions''. We first review \cite{aleks2023subspace}'s technical notion of what an ``interpretability illusion'' is, and then we show that even intuitive and desirable explanations can qualify as illusions in this sense. As a result, their method of discovering ``illusions'' can reject explanations they consider ``non-illusory''. We then argue that the illusions \citet{aleks2023subspace} see in practice are artifacts of their training and evaluation paradigms. We close by emphasizing that, though we disagree with their core characterization, \citet{aleks2023subspace}'s examples and discussion have undoubtedly pushed the field of interpretability forward.

\end{abstract}

\section{Introduction}

It has long been known that individual neurons in trained neural networks often play multiple roles \citep{Smolensky1986,PDP1,PDP2}, giving rise to what \citet{Smolensky1986} calls “patterns”: distributed representations defined by linear combinations of unit vectors. In the recent wave of interpretability research, such distributed representations have once again entered the limelight under the guise of ``polysemantic neurons'' \citep{olah2020zoom, bricken2023monosemanticity}. Furthermore, the widely discussed ``linear representation hypothesis'' \citep{mikolov-etal-2013-linguistic,elhage2022toy,nanda-etal-2023-emergent,park2023linear} takes the fundamental unit of analysis to be linear subspaces of neural activations.

It is vital that our model interpretability methods have the capacity to find this kind of structure if it does exist. Not all recent methods achieve this, however. In particular, the classical interchange interventions of \citet{Geiger-etal-2020} (also known as activation patching) presuppose that neurons will play unique causal roles~\citep{vig2020causal,Csordas2021,CausalLM,Ravfogel:2020,Elazar-etal-2020,de2021sparse,abraham-etal-2022-cebab,chan2022causal,wang2022interpretability} %
Thus, these methods will inevitably fail to uncover the more abstract representations that \citet{PDP2}, \citet{Smolensky1986}, and others sought to characterize.

In response to this limitation of interchange interventions, we recently proposed \emph{distributed interchange interventions}, which intervene on subspaces and can be thought of as remapping sets of neurons in the original network using a learned change-of-basis matrix, and we developed Distributed Alignment Search (DAS) as a method for finding this matrix \citep{geiger2023finding}. Our experiments with DAS showed that these interventions can reveal previously overlooked aspects of the causal structure of networks trained to solve hard problems. For instance, a simple network solving a hierarchical equality task was found to carry out an intuitive symbolic computation exactly, but only in distributed representations \citep{geiger2023finding}.

In a recent paper, \citet{aleks2023subspace} argue that distributed interchange interventions can give rise to what they call ``interpretability illusions'': situations in which neurons that are causally disconnected in a normal run of the model come to play causally efficacious roles when distributed interchange interventions are performed. An earlier version of the argument is given in a blog post on LessWrong \citep{lange2023illusion}.

Here we respond to \citeauthor{aleks2023subspace}'s argument. Fundamentally, we reject the label ``illusions'' for these phenomena. These are simply discoveries about distributed representations. The crux of our response is a technical observation: it follows from \citeauthor{aleks2023subspace}'s definition that a so-called illusion arises where (and essentially only where) the distributed interchange intervention leads the model to induce representations that are not orthogonal to the nullspace of the model’s subcomponents (e.g., nullspace of the down-projection weights in the MLP layer of GPT-2). There is nothing problematic about such situations, though, and indeed we show that these situations arise with standard interchange interventions or activation patches. Overall, then, we should not dismiss these phenomena as illusions. They are simply facts about how networks work, and our interpretability methods should be able to discover them. Indeed, as we describe later, the prevalence of such situations reflects the fact that representations will reflect the natural variation in activations driven by inputs, and this variation need not vanish in null directions. 

In the present paper, we first review \citeauthor{aleks2023subspace}'s technical notion of what an ``interpretability illusion'' is, and then we show that even intuitive and desirable explanations termed ``non-illusory'' by \citeauthor{aleks2023subspace} can qualify as illusions in this sense. The fundamental insight here is the general one we identified above: unless the inputs and all subsequent representations are orthogonal to the nullspace of the relevant model components (unlikely in practice, and not something we would impose), so-called illusions are inevitable. We then argue that the illusions \citeauthor{aleks2023subspace} see in practice are artifacts of their training and evaluation paradigms. Furthermore, we conduct additional analyses on the indirect object identification (IOI) task, aiming to reproduce the findings of \citeauthor{aleks2023subspace} and to offer further insights.
We close by emphasizing that, though we disagree with their core characterization, \citeauthor{aleks2023subspace}'s examples and discussion have undoubtedly pushed the field of mechanistic interpretability forward.

\section{Background: Defining \citet{aleks2023subspace}'s ``Illusion''}\label{Sec:illusion-definition}
We first formally distill \citet{aleks2023subspace}'s definition of ``illusion''. 

\subsection{Set-up}
The main objective of DAS is to intervene on activations in a subspace represented by a set of orthonormal vectors $v \in \mathbb{R}^{n\times d}$ where each vector has $d$ elements (corresponding to the dimension of the locations where we’ve chosen to look for a representation). Let’s take two examples from our data distribution, $\mathbf{A}$ and $\mathbf{B}$. Intervening on activations created by $\mathbf{A}$ with $\mathbf{B}$ in the subspace spanned by $v$ can be expressed as,
\begin{equation}\label{Eqn:DAS}
u_{\mathbf{A}}^{v \leftarrow \mathbf{B}} = u_{\mathbf{A}} + (u_{\mathbf{B}}v^\intercal - u_{\mathbf{A}}v^\intercal)v
\end{equation}
where $u_{\mathbf{A}}, u_{\mathbf{B}} \in \mathbb{R}^{d}$ are the original activations in the intervention location, using $\mathbf{A}$ (resp.~$\mathbf{B}$) as the input to the model.

\subsection{Nullspace Decomposition}
Specializing to the case where the locations of interest are just prior to the down-projection layer of an MLP, \citeauthor{aleks2023subspace} decompose $v$ into two orthogonal parts as,
\begin{equation}\label{Eqn:decompose}
v = v_{\text{nullspace}} + v_{\text{rowspace}}
\end{equation}
where the definitions of these two sub-components are,
\begin{itemize}
  \item $v_{\text{nullspace}}$: $v$'s projection onto the nullspace of $W_{\text{out}} \in \mathbb{R}^{d\times o}$, the weights of the downward projection layer in the MLP with an input dimension of $d$ and an output dimension of $o$. 
  \item $v_{\text{rowspace}} = v - v_{\text{nullspace}}$: the complement.
\end{itemize}

\subsection{The ``Illusion''}
The \emph{intervened output} of the MLP layer as the result of intervention along $v$ in its input can be written out based off \Eqnref{Eqn:DAS},
\begin{equation}
u_{\mathbf{A}}^{v \leftarrow \mathbf{B}}W_{\text{out}} = u_{\mathbf{A}}W_{\text{out}} + (u_{\mathbf{B}}v^\intercal - u_{\mathbf{A}}v^\intercal)vW_{\text{out}}
\end{equation}
Now, let’s expand the above equation with the decomposition provided in \Eqnref{Eqn:decompose},
\begin{equation}
u_{\mathbf{A}}^{v \leftarrow \mathbf{B}}W_{\text{out}} = u_{\mathbf{A}}W_{\text{out}} + (u_{\mathbf{B}}v_{\text{n}}^\intercal + u_{\mathbf{B}}v_{\text{r}}^\intercal - u_{\mathbf{A}}v_{\text{n}}^\intercal - u_{\mathbf{A}}v_{\text{r}}^\intercal)(v_{\text{n}}+v_{\text{r}})W_{\text{out}}
\end{equation}
where we denote $v_{\text{nullspace}}$ as $v_{\text{n}}$, and $v_{\text{rowspace}}$ as $v_{\text{r}}$. Now, let’s reorganize things a little and rewrite the right-hand side (RHS) as,
\begin{equation}
u_{\mathbf{A}}^{v \leftarrow \mathbf{B}}W_{\text{out}} = u_{\mathbf{A}}W_{\text{out}} + ((u_{\mathbf{B}} - u_{\mathbf{A}})v_{\text{n}}^\intercal + (u_{\mathbf{B}} - u_{\mathbf{A}})v_{\text{r}}^\intercal)(v_{\text{n}}+v_{\text{r}})W_{\text{out}}
\end{equation}
and expand the RHS to five different parts,
\begin{equation}
\begin{split}
u_{\mathbf{A}}^{v \leftarrow \mathbf{B}}W_{\text{out}} & = u_{\mathbf{A}}W_{\text{out}} + (u_{\mathbf{B}} - u_{\mathbf{A}})v_{\text{n}}^\intercal v_{\text{n}}W_{\text{out}} \\
 & + (u_{\mathbf{B}} - u_{\mathbf{A}})v_{\text{r}}^\intercal v_{\text{n}}W_{\text{out}} \\
 & + (u_{\mathbf{B}} - u_{\mathbf{A}})v_{\text{n}}^\intercal v_{\text{r}}W_{\text{out}} \\
 & + (u_{\mathbf{B}} - u_{\mathbf{A}})v_{\text{r}}^\intercal v_{\text{r}}W_{\text{out}}
\end{split}
\end{equation}
We know that $xv_{\text{n}}W_{\text{out}} = 0$, for any vector $x$. Thus, we can cross out the first two terms after $u_{\mathbf{A}}W_{\text{out}}$ as they are always 0. The RHS thus can be simplified as,
\begin{equation}\label{Eqn:illusion}
\begin{split}
u_{\mathbf{A}}^{v \leftarrow \mathbf{B}}W_{\text{out}} & = u_{\mathbf{A}}W_{\text{out}} + (u_{\mathbf{B}} - u_{\mathbf{A}})v_{\text{n}}^\intercal v_{\text{r}}W_{\text{out}} \\
 & + (u_{\mathbf{B}} - u_{\mathbf{A}})v_{\text{r}}^\intercal v_{\text{r}}W_{\text{out}}
\end{split}
\end{equation}
The equation above shows that the causal effects of the subspace interventions can be written in  two parts: (1) the projected difference between two examples onto the \emph{nullspace} of $W_{\text{out}}$, and (2) the projected difference between two examples onto the \emph{rowspace} of $W_{\text{out}}$.

\citeauthor{aleks2023subspace}'s intuition about ``interpretability illusion'' is that variations of activations along $v_{\text{n}}$ (i.e., the \emph{nullspace} of $W_{\text{out}}$) should not have any causal effect on model predictions. Thus, the causal effects of intervening along $v$ and $v_{\text{r}}$ are approximately the same for any non-illusory $v$. Following this intuition, the main experiments that \citeauthor{aleks2023subspace} designed is to compare the causal effect on model predictions of intervening on $v$ with intervening on $v_{\text{r}}$ alone (as noted in Section 3.4 and the experimental sections of the paper). Formally, the ``interpretability illusion'' of intervening on the targeted MLP layer can be defined as,
\begin{equation}\label{Eqn:illusion-effect}
\textsc{IllusionEffect}(\Phi, W_{\text{out}}, v, \mathbf{A}, \mathbf{B}) = \Phi(u_{\mathbf{A}}^{v \leftarrow \mathbf{B}}W_{\text{out}}) - \Phi(u_{\mathbf{A}}^{v_{\text{r}} \leftarrow \mathbf{B}}W_{\text{out}})
\end{equation}
where the model’s downstream components are folded into $\Phi(\cdot)$. Assuming intervening on $v$ produces the desired counterfactual behavior in general, an ``illusion'' arises if the causal effect of intervening on $v_{\text{r}}$ is \emph{much smaller} than $v$ towards the desired goal (e.g., flip the model’s prediction from the correct name to the incorrect name in their indirect object identification (IOI) experiment with GPT-2).

\section{Revisiting \citet{aleks2023subspace}'s Toy Example}
\citet{aleks2023subspace} introduce an illustrative toy example. We revisit this example in some detail before returning to more complex models.

\subsection{Set-up}
The toy example involves a simple linear neural network $f(x)=(xW_1^\intercal)W_2$ where $W_1 = [1, 0, 1]^\intercal$ and $W_2 = [0, 2, 1]^\intercal$. As a result, this network implements a ``copy'' function as,
\begin{equation}
f(x) = 0\times(1\times x)_{\text{H1}} + 2\times(0\times x)_{\text{H2}} + 1\times(1\times x)_{\text{H3}} = x
\end{equation}
where we use $(*)_{\text{H1}}$, $(*)_{\text{H2}}$ and $(*)_{\text{H3}}$ to index the activations of three hidden representations. We can then write out the nullspace of $W_2$ as a plane linearly spanned by two vectors, $[1, 0, 0]$ and $[0, -\frac{1}{\sqrt{3}}, \frac{2}{\sqrt{3}}]$ (i.e., a trivial vector where the weighted sum of last two dimensions nullify the causal effect). The other relevant geometric structure is what we will call the \textbf{data-induced submanifold}, which is the subspace of activations that can be achieved by some input to the network. It is clear in this simple example that the data-induced submanifold is the line extended by the unit vector $[\frac{1}{\sqrt{2}}, 0, \frac{1}{\sqrt{2}}]$,  the image of $W_1$.

In the toy example, the $\textsc{IllusionEffect}$ definition in \Eqnref{Eqn:illusion-effect} can be further simplified given \Eqnref{Eqn:illusion} and the fact that there are no downstream computations (i.e., $\Phi_{\text{toy}}$ in \Eqnref{Eqn:illusion-effect} is a no-op) as,
\begin{equation}\label{Eqn:illusion-effect-toy}
\textsc{IllusionEffect}(\Phi_{\text{no-op}}, W_{2}, v, \mathbf{A}, \mathbf{B}) = u_{\mathbf{A}}^{v \leftarrow \mathbf{B}}W_{2} - u_{\mathbf{A}}^{v_{\text{r} \leftarrow \mathbf{B}}}W_{2} = (u_{\mathbf{B}} - u_{\mathbf{A}})v_{\text{n}}^\intercal v_{\text{r}}W_{2}
\end{equation}

\subsection{An Obvious Non-Illusory Direction}
In the toy example, activations on $\text{H3}$ are taken to mediate the signal through the network. \citeauthor{aleks2023subspace} are worried that interchange interventions along directions like $[\frac{1}{\sqrt{2}}, \frac{1}{\sqrt{2}}, 0]$ would nullify $\text{H3}$ and create a new causal pathway by intervening on the sum of $\text{H1}$ and $\text{H2}$. By contrast, \citeauthor{aleks2023subspace} identify $v^{\text{non-illusory}} = [0, 0, 1]$ as an obvious non-illusory direction in this case since it identifies $\text{H3}$ as the only causally relevant activation.

Intriguingly, we can see that the direction $v^{\text{non-illusory}}$ is \textbf{not} orthogonal to the nullspace of $W_2$ by simply checking the dot-product between $v^{\text{non-illusory}}$ and any vector on the nullspace of $W_2$ such as $[0, -\frac{1}{\sqrt{3}}, \frac{2}{\sqrt{3}}]$. This entails that the non-illusory representation will have an illusion effect as defined in \Eqnref{Eqn:illusion-effect-toy}. To see this in detail we work through the calculations. We can decompose $v^{\text{non-illusory}}$ into two non-zero orthogonal parts following \citeauthor{aleks2023subspace}'s paradigm as,
\begin{equation}
v^{\text{non-illusory}} = v_{\text{n}}^{\text{non-illusory}} + v_{\text{r}}^{\text{non-illusory}}
\end{equation}
where we denote $v_{\text{nullspace}}^{\text{non-illusory}}$ as $v_{\text{n}}^{\text{non-illusory}}$, and $v_{\text{rowspace}}^{\text{non-illusory}}$ as $v_{\text{r}}^{\text{non-illusory}}$. To find $v_{\text{n}}^{\text{non-illusory}}$ as the orthogonal projection of $v^{\text{non-illusory}}$ onto the nullspace pf $W_2$, we create a matrix $\mathbf{M} \in \mathbb{R}^{3\times 2}$ representing the span of two vectors in the nullspace of $W_2$ as,
\[
\mathbf{M}
= 
\begin{bmatrix}
    1 & 0\\
    0 & -\frac{1}{\sqrt{3}}\\
    0 & \frac{2}{\sqrt{3}}\\
\end{bmatrix}
\]
We can then get the orthogonal projection of $v^{\text{non-illusory}}$ onto this plane as, 
\begin{equation}
v_{n}^{\text{non-illusory}} = \text{proj}_{\mathbf{M}}v^{\text{non-illusory}} = \mathbf{M}(\mathbf{M}^\intercal \mathbf{M})^{-1}\mathbf{M}^\intercal v^{\text{non-illusory}}
\end{equation}
As a result, we get,
\begin{equation}
v_{n}^{\text{non-illusory}} = [0, -0.4, 0.8]
\end{equation}
\begin{equation}
v_{r}^{\text{non-illusory}} = v - v_{n}^{\text{non-illusory}} = [0, 0, 1] - [0, -0.4, 0.8] = [0, 0.4, 0.2]
\end{equation}
To derive the exact $\textsc{IllusionEffect}$ term, we can sample two random inputs $\mathbf{A}$ (as example $x$) and $\mathbf{B}$ (as example $x'$) to first obtain hidden representations realized by our simple network as,
\[
u_{\mathbf{A}} = [x, 0, x]; u_{\mathbf{B}} = [x', 0, x']; u_{\mathbf{B}} - u_{\mathbf{A}} = [x' - x, 0, x' - x]
\]
Now, $\textsc{IllusionEffect}$ as in \Eqnref{Eqn:illusion-effect-toy} can be calculated step-by-step as,
\begin{equation}
\begin{split}
\textsc{IllusionEffect}(\Phi_{{\text{no-op}}}, W_{\text{out}}, v, \mathbf{A}, \mathbf{B}) & = (u_{\mathbf{B}} - u_{\mathbf{A}})v_{\text{n}}^\intercal v_{\text{r}}W_{2} \\
& = [x' - x, 0, x' - x][0, -0.4, 0.8]^\intercal [0, 0.4, 0.2][0, 2, 1]^\intercal \\
& = 0.8(x' - x)[0, 0.4, 0.2][0, 2, 1]^\intercal \\
& = (x' - x)[0, 0.32, 0.16][0, 2, 1]^\intercal\\
& = 0.8(x' - x)
\end{split}
\end{equation}
Recall that $\textsc{IllusionEffect}$ represents the difference in causal effect between intervening on $v$ and $v_r$ alone. This shows the causal effect by intervening with $v_r$ alone \emph{greatly diminishes} by only retaining 20\% of the causal effect when intervening on $v$. For instance, when $x = 1$ and $x' = 5$, intervening along $v$ will result in an output of 5, and intervening along $v_r$ will result in an output of $5 - 0.8\times(5-1) = 1.8$ which is \emph{much lower}.\footnote{\citeauthor{aleks2023subspace}'s original implementation also normalizes $v_{n}^{\text{non-illusory}}$ and $v_{r}^{\text{non-illusory}}$, which will lead to even larger ``illusory'' effect, given that the norms of both vectors are smaller than 1, thereby further invalidating this non-illusory direction.} \citeauthor{aleks2023subspace} identify $v_{n}^{\text{non-illusory}}$ as an obvious non-illusory direction in the toy example. However, we are able to show that $v_{n}^{\text{non-illusory}}$ would be classified as an ``illusory'' direction according to their own definitions (outlined in Section 3.4).

\subsection{A Broader Lesson from the Example above}
We have shown that \citet{aleks2023subspace}'s paradigm can reject a non-illusory direction. More importantly, we also have shown that a faithful distributed representation may \textbf{not} be orthogonal to the nullspace of downstream computation. \textbf{Why?} Simply because the image of upstream computations need not be (and indeed generically will not be). To expand that: the space of possible inputs to the network leads to variation within a given set of neurons that covers some data-induced submanifold. This submanifold need not, and generally will not, span the whole activation space. It should be clear that, if our goal is understanding how the network computes, we should be interested in variation along this submanifold; indeed interchange interventions will operate on this submanifold by construction. However, there is no reason to insist that the nullspace of downstream computations must be orthogonal to this data-induced submanifold. Hence we agree with \citeauthor{aleks2023subspace} that, for example, their MLP-in-the-middle ``illusion'' will be common (see Section 7 in the paper). Except it is not an ``illusion'' in any useful sense. It is a simple fact about the geometry of representations. We believe the work of \citet{aleks2023subspace} highlights the need to think carefully about these geometric relationships and can lead to important research directions. 

Beyond questions of ``illusory'' causation, the toy example provides a lovely case of multiple abstraction. In brief, a computation such as $h(x)=x-x+x$ can be seen as implementing the identity by copying the input in \emph{two different ways}.
As we describe in \Appref{App:multi-abstractions}, the toy example is an instance of this pattern with one copy of a distributed representation.
The possibility of such multiple abstractions is connected with the fundamentals of the causal abstraction framework and the lack of preferred bases in linear algebra.

\section{Remarks on \citet{aleks2023subspace}'s Experimental Evidence for Discovering ``Illusions'' in the Wild}

We close by reviewing the experimental evidence that \citet{aleks2023subspace} offer for the claim that their ``illusions'' arise in real pretrained LMs. Our view is that the evidence provided in favor of this claim is not strong. However, for the reasons given above, we would find substantiated examples of this effect to be interesting evidence of latent causal structure rather than a worrisome illusion.

\subsection{Background: the indirect objective identification (IOI) and factual recall tasks}
\citet{aleks2023subspace} conduct two experiments to find ``illusions'' in the wild. We first describe their two tasks in detail.
\paragraph{Indirect Objective Identification (IOI)} asks the model to complete sentences of the form ``When Mary and John went to the store, John gave a bottle of milk to'' with the correct answer being ``Mary''. ``John'' is the repeated name as S (i.e., the subject), and ``Mary'' is the non-repeated name as IO (i.e., the indirect object). GPT-2 is effective at solving this task in a zero-shot fashion \citep{wang2022interpretability}. \Figref{Fig:IOI-experiment} shows one high-level causal model that can solve this task, and the goal is to find alignments between activations and high-level causal variables (e.g., the name position variable or the correct IO name variable).

\paragraph{Factual Recall} involves subject-relation-object triple $(s,r,o)$ (e.g., $s=$``Eiffel Tower'', $r=$``is in'', $o=$``Paris''). We test whether a model can recall the fact associated with a triple by constructing a prompt using a subject-relation tuple (e.g., ``The Eiffel Tower is in'') and checking whether the model can generate the correct output (e.g., ``Paris'') given the prompt. Given any pair of triples that a model can recall correctly, $\textbf{A}=(s,r,o); \textbf{B}=(s',r,o')$, the goal is to intervene on the model's activations induced using $\textbf{A}$ with activations using $\textbf{B}$ and have the output changed from $o$ to $o'$.

\subsection{Interchange Intervention Accuracy (IIA)}
In our work on causal abstraction (starting from the initial work of \citealt{Geiger-etal-2020}), we evaluate the degree to which a high-level model is a faithful description of a neural network based on our ability to control model behavior through interventions on neural activations aligned with high-level variables. Interchange intervention accuracy (IIA) is the percentage of counterfactual predictions matched under interchange interventions between the high-level causal model and the neural model. Although IIA has limitations (as outlined in \citealt{geiger2023finding,wu-etal-2023-Boundless-DAS}), it offers an intuitive measure for gauging the interpretability of a neural model in the following sense: IIA is bounded by 100\%, and when IIA is 100\%, the causal model is a faithful explanation of how the neural network behaves under the set of interventions we perform, which provides transparent insights into the neural network’s causal mechanisms \citep{geiger-etal-2021-iit}. In practice, achieving 100\% IIA is uncommon, but the higher the value of IIA, the more confidence we have in reasoning about the high-level causal model as a stand-in for the low-level neural model.

In \citet{aleks2023subspace}'s IOI experiment, they try to learn a single DAS direction that abstracts the name position information. For instance, if the IO (i.e., correct output token) name appears as the second name in the base, and the first name in the source example, the counterfactual label for the base would switch to the first name mentioned in the base after the intervention. When finding such directions, \citet{aleks2023subspace} end up with an IIA ranging from 0.0\% to 4.2\% for intervention locations in MLP layers, as shown in their Table 1. This is far from a strong signal given the fact that GPT-2 can achieve a task performance of 91.6\%.\footnote{\citet{aleks2023subspace} reported 91.6\% accuracy in their publicly released codebase. We reproduced this number and provided detailed scripts in \url{https://github.com/frankaging/pyvene/blob/main/tutorials/advanced_tutorials/IOI_with_DAS.ipynb}.} In our work, we routinely see significantly higher IIA values. For example, in the study of how Alpaca solves a simple reasoning task in context \citep{wu-etal-2023-Boundless-DAS}, we see IIA values around 85\% even in ambitious, out-of-domain evaluations. Thus, we interpret 4\%, which is much lower than the task performance, as a failure to find any relevant structure.

Instead of relying on IIA, \citet{aleks2023subspace} use a new metric, which they call the fractional logit difference decrease (FLDD) to quantify the actual ``illusion'' effect. FLDD measures a fractional logit difference between the IO and S (i.e., incorrect output token) tokens before and after the intervention, as described in Equation 4 in the paper. This measure is interesting because it encodes a qualitative pattern that very roughly tracks IIA. Assuming near-perfect performance of the model on a given task, when FLDD is above 100\% this means the intervention, on average, produces the right counterfactual prediction (e.g., switching from IO to S in the IOI experiment). As their Table~1 shows, having FLDD greater than 100\% does correlate somewhat with relatively high IIA scores.

However, the measure may also mislead one into thinking that the magnitudes themselves are closely reflective of counterfactual model behavior. Because FLDD is based on a ratio of log-likelihoods, it can be brought arbitrarily close to 100\% \textbf{with no effect whatsoever on counterfactual behavior}. For instance, if the model predicts IO with a probability of 0.505, while the intervention leads to a slightly lower probability of $0.5025$, this would mean approximately 50\% FLDD, even though model behavior does not change at all. This is why their Table 1 reveals a generally poor correlation with IIA (e.g., 46.7\% FLDD results in only 4.2\% IIA; see also their discussion on p.~14).

Despite the limitations of FLDD, we are enthusiastic about considering other metrics than IIA for evaluating model explanations, including those that compare logit differences (e.g., KL divergence).

\subsection{Checking Dormant or Disconnected Components via Correlational Analysis of Activations}

\citet{aleks2023subspace} extensively study the distributions of projected activations when projecting onto nullspace and rowspace to further validate their hypothesis around dormant or disconnected pathways in GPT-2. Furthermore, they often draw conclusions by analyzing the activation distributions. For instance in their Figure 5, they observe that activation projections onto the nullspace separate more saliently across two groups compared to activation projection onto the rowspace. On this basis, they claim that the rowspace is predominantly dormant and the nullspace projection drives the causal effect. 

While it is interesting to observe how activations are distributed differently, we want to point out that they may carry less causally interesting information than one might expect. For example, \citet{huang2023assess} assess the proposal of \citet{bills2023language} to use GPT-4 to explain neurons in \mbox{GPT-2-XL}. One of their core observations is that activations that are found to be strongly correlated with a certain high-level concept can have little to no causal effect on model predictions in downstream tasks where the concept is being used. We thus argue that analysis solely based on activations cannot support strong causal claims about model structures.

\subsection{Potential Pitfalls in the Factual Recall Experiment}

\citet{aleks2023subspace} also try to replicate their finding of ``illusion'' in the factual recall task as in their Section 6.1 (p. 20). 
Their core experimental pipeline for finding ``illusions'' relies on learning a single DAS direction for a single pair of facts with MLP activations at different layers of GPT2-XL.\footnote{They compare the causal effect between intervening on learned DAS direction $v$ and $v_{\text{rowspace}}$ (i.e., the rowspace decomposition as shown in \Secref{Sec:illusion-definition}) for the seen pair of facts to trace ``illusion''. Details can be found in \citet{aleks2023subspace}'s original implementation at \url{https://github.com/amakelov/activation-patching-illusion/blob/main/fact_patching.ipynb}.} In other words, \citet{aleks2023subspace} find a DAS direction that would change the label of the base example to the source example when intervening on that single DAS direction. This is \textbf{not} an experimental design that we would endorse. Two potential pitfalls are worth calling out for future research when using DAS to find alignments between neural networks and high-level causal models.

First, the learned DAS direction aligns the neural network with an uninteresting high-level causal model: a dummy model that always returns the answer of the seen source example. It should be radically unsurprising to find the ``return the answer'' high-level causal model is indeed a causal abstraction of any model that returns the answer. Consequently, we suspect there could be many DAS directions that could achieve this alignment with a seemingly perfect alignment score (as shown in Figure 7 on p. 20), but such results are not meaningful explanations. Second, the current experimental setup also entails that the learned DAS direction overfits the single pair of facts seen during training. DAS is a model with learned parameters. As such (as with essentially all ML models), it should be trained on numerous examples and tested on examples that are disjoint from the train set. \citet{aleks2023subspace}'s factual recall experiment violates both of these conditions. As a result, their directions would likely not generalize to any other facts except the ones it is trained on. Additionally, the latter point is related to the former: generalization would show evidence of alignment with a more interesting high-level model.

\begin{figure*}[tp]
  \centering
  \includegraphics[width=0.95\linewidth]{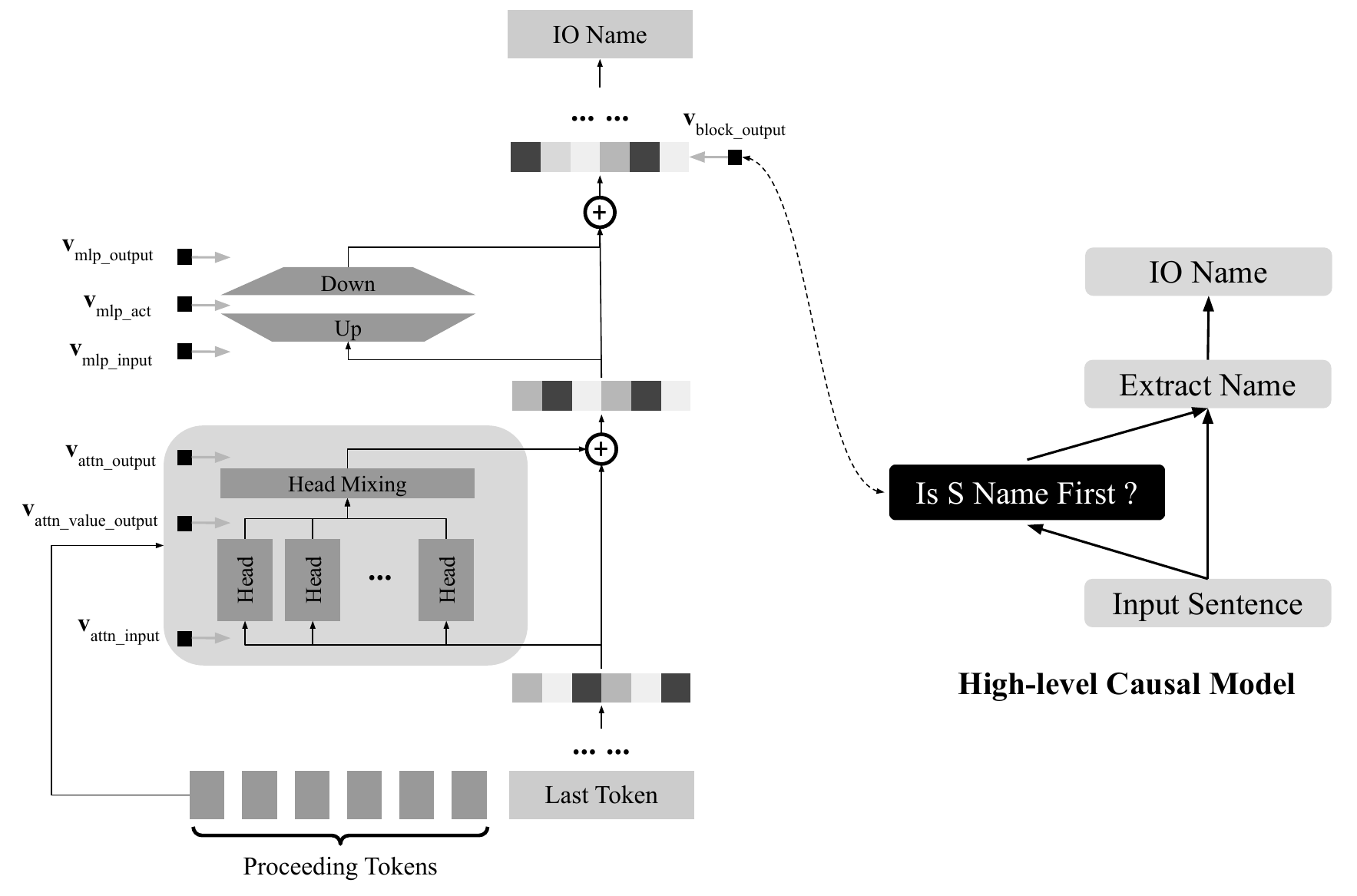}
  \caption{An illustration of aligning a high-level causal model with key intervention locations of the streams on top of the last input token in the GPT-2 model. Besides aligning the main residual streams and the MLP activations, we align other streams to study how the name position information emerges in the GPT-2 model. The head mixing layer is a linear layer.}
  \label{Fig:IOI-experiment}
\end{figure*}

\section{Additional Indirect Object Identification (IOI) Results with GPT-2}\label{Sec:results-in-IOI}
In this section, we show that DAS can be productively used to reproduce some of the key findings outlined in previous mechanistic interpretability results on the IOI task~\citep{wang2022interpretability, aleks2023subspace}. Furthermore, we show how to use DAS to gain new insights into causal mechanisms induced by GPT-2 on IOI.\footnote{To foster reproducibility,  we release our code and data at \url{https://github.com/frankaging/pyvene/blob/main/tutorials/advanced_tutorials/IOI_with_DAS.ipynb}.}

\subsection{Set-up}
We use the same model and dataset generation script as \citet{wang2022interpretability} and \citet{aleks2023subspace}. Following \citet{aleks2023subspace}, we learn a single DAS direction to localize the name position in the IOI task. Since there are only two names (i.e., the S or IO name) mentioned in the input sentence, the name position information can be binarized (e.g., a binary variable indicating whether S is the first name or not) and thus represented in a single dimension. As shown in \Figref{Fig:IOI-experiment}, one possible high-level causal model of solving the IOI task is to first determine whether the S name is the first name mentioned in the sentence, and then extract the name based on the intermediate result of the first step before returning the final answer. In our experiments as well as \citet{aleks2023subspace}'s experiments, DAS is used to find causal variables representing the name position information (i.e., a causal variable representing whether S name is the first name or not as shown in \Figref{Fig:IOI-experiment}). We try seven intervention locations to find the causal variable as annotated in \Figref{Fig:IOI-experiment}. We use IIA as our evaluation metric. Other details about our experimental set-up can be found in \Appref{App:experiment-setup}.

\subsection{The Name Position Information in the Main Residual Stream and the MLP Activations}
\begin{figure*}[t]
    \centering
    \begin{minipage}{0.48\textwidth}
        \centering
        \includegraphics[width=\textwidth]{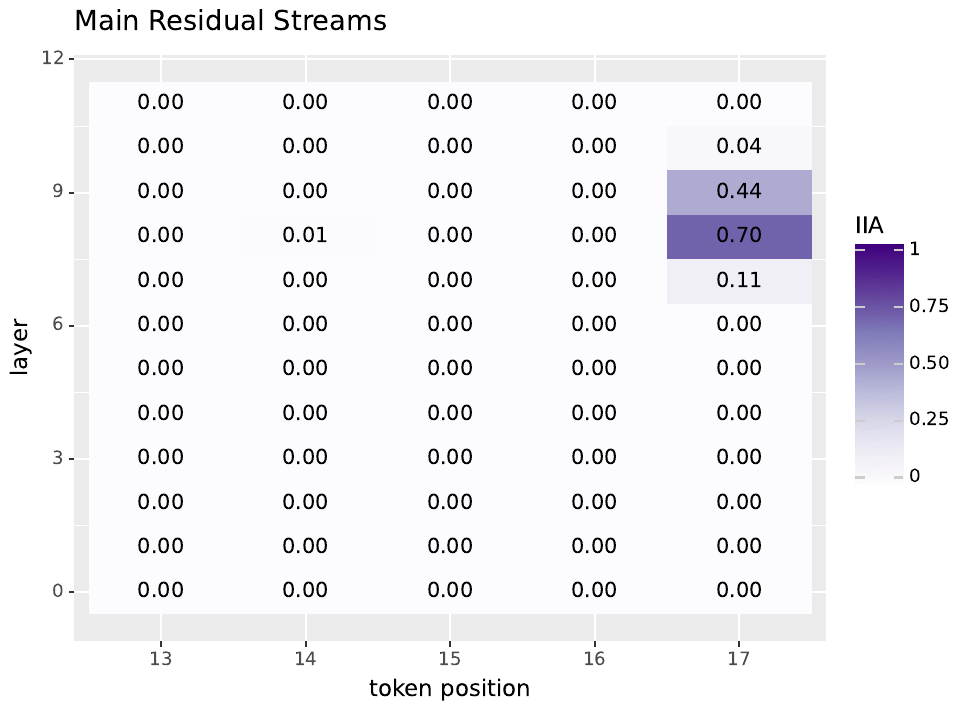}
    \end{minipage}
    \hfill %
    \begin{minipage}{0.48\textwidth}
        \centering
        \includegraphics[width=\textwidth]{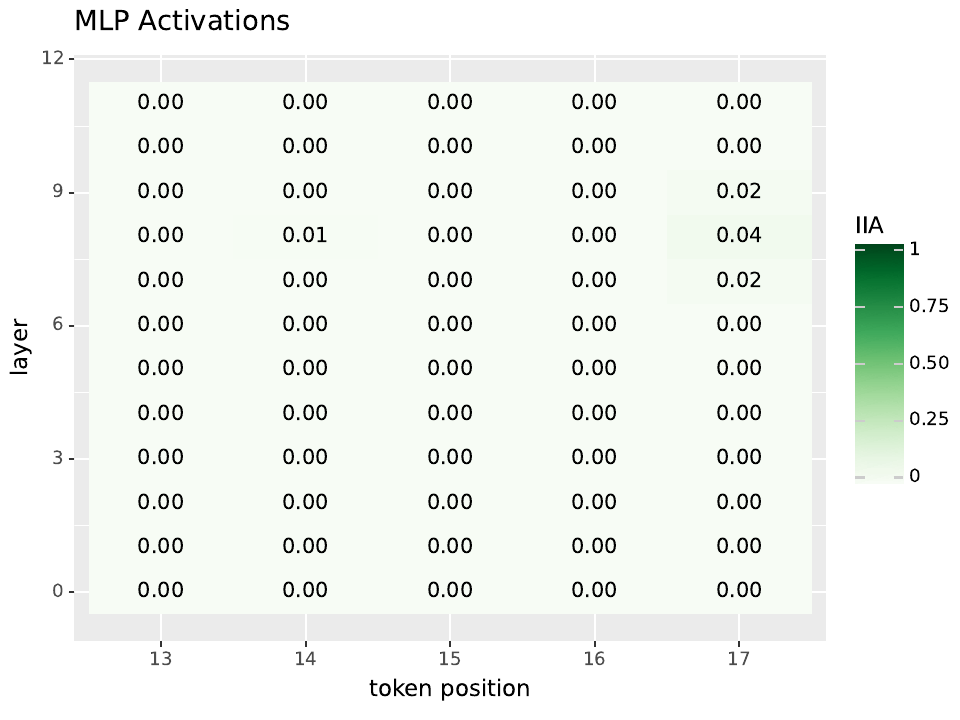}
    \end{minipage}
    
    \caption{Interchange Intervention Accuracy (IIA) when aligning the name position variable with different intervention locations in the main residual streams ($v_{\text{block\_out}}$) as well as the MLP activations ($v_{\text{mlp\_act}}$) above the last token and the four tokens proceeding it. The GPT-2 model achieves 96\% task accuracy. Higher IIA means better alignment. Overall our results are consistent with \citet{aleks2023subspace}'s findings where name position information mainly resides above the last token at the 8th layer.}
    \label{Fig:residual_and_mlp_act}
\end{figure*}

As shown in \Figref{Fig:residual_and_mlp_act}, our results are consistent with \citet{aleks2023subspace}'s findings, where the name position information best aligns with the 8th\footnote{Author correction: we use 0-indexing that ``0th'' layer means the first layer.} layer in GPT-2. The learned DAS direction achieves an IIA of 70\% with the main residual stream ($v_{\text{block\_out}}$) and an IIA of 4\% with the MLP activations ($v_{\text{mlp\_act}}$), which is approximately the same as \citeauthor{aleks2023subspace}'s results (see their Table~1). Our IIA scores suggest that MLP activations carry hardly any information about name position information.

Our results also suggest that single token representations before the last token rarely carry any name position information.\footnote{We additionally align with each token representation other than the four proceeding tokens before the last token and find IIAs are approximately zero.} Furthermore, we use DAS to check whether name position information is distributed across tokens by learning a single DAS direction on the concatenated token representation of all the tokens before the last token (16 tokens in total) at the 7th layer (a layer before we find name position information). We find an IIA of 3\%, which means name position information is not distributed across tokens either. This gives another piece of evidence that name position information emerges at the 8th layer only.  

\begin{figure*}[h]
    \centering
    \begin{minipage}{1.0\textwidth}
        \centering
        \includegraphics[width=\textwidth]{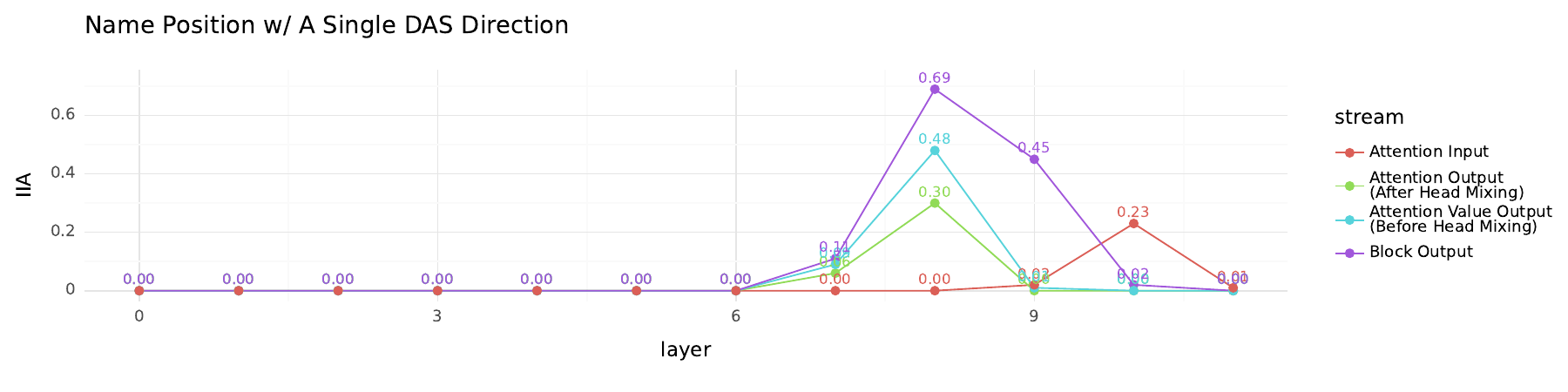}
    \end{minipage}
    \begin{minipage}{1.0\textwidth}
        \centering
        \includegraphics[width=\textwidth]{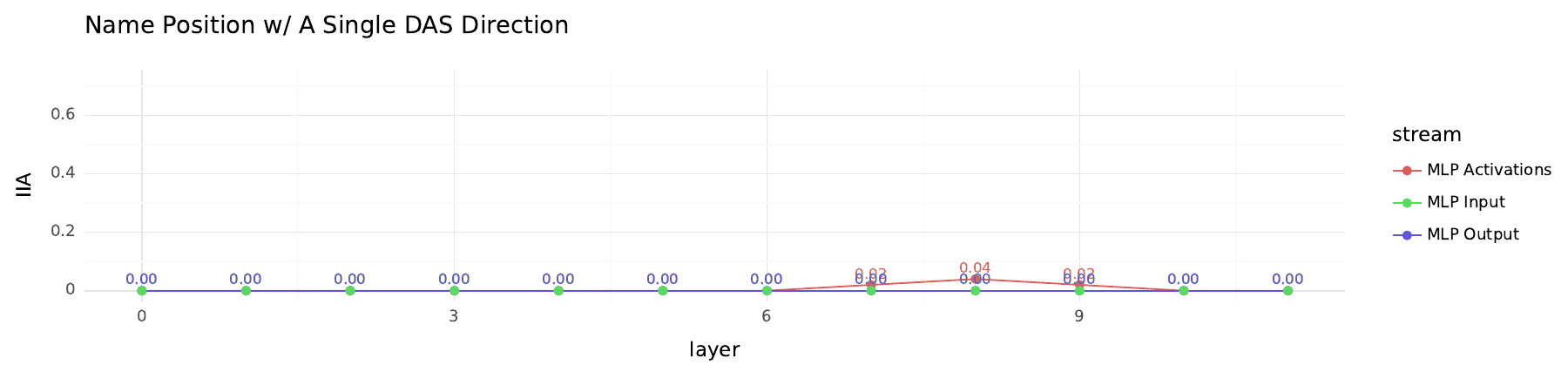}
    \end{minipage}
    \caption{Interchange Intervention Accuracy (IIA) when aligning the name position variable with different intervention locations above the last token.}
    \label{Fig:last_token_streams}
\end{figure*}

\subsection{Different Streams Show Where the Name Position Information Emerges}
As shown in \Figref{Fig:IOI-experiment}, we align with seven different locations across layers above the last token to trace the name information across components. As shown in \Figref{Fig:last_token_streams}, consistent with previous work~\citep{wang2022interpretability, aleks2023subspace}, the MLP layers carry little to no information about the name position. For instance, the MLP input and output streams ($v_{\text{mlp\_input}}$ and $v_{\text{mlp\_output}}$) both flatten out at 0\% IIA across all layers. 

Our results also suggest that the name position information emerges most saliently through the self-attention block at the 8th layer. As shown on the top panel of \Figref{Fig:last_token_streams}, the block input stream ($v_{\text{block\_input}}$) of the 8th layer (i.e., the block output stream of the previous layer) reaches about 11\% IIA, the attention output stream ($v_{\text{attn\_out}}$) reaches about 48\% IIA, and the resulting block output stream reaches about 69\% IIA. These results suggest that the self-attention block is largely responsible for adding the name position information into the residual stream. Furthermore, a similar trend also happens in the 7th layer, suggesting the name information emerges across layers in a distributed way as well. Strikingly, the self-attention streams after the 8th layer carry nothing about the name position information suggesting these self-attention heads can be pruned, as shown in previous work~\citep{wang2022interpretability}. 

DAS assumes a fixed-size linear subspace; however, a binary variable may not always fit within a linear subspace of a single dimension. Thus, we also experimented with Boundless DAS \citep{wu-etal-2023-Boundless-DAS}, a variant of DAS that does not assume a fixed dimensionality for the DAS subspace. With Boundless DAS, the boundaries of the distributed intervention are learned along with the change-of-basis matrix. Supporting results using this method can be found in \Appref{App:additional_boundless_das_analyses}.

\begin{figure*}[t]
    \centering
    \begin{minipage}{1.0\textwidth}
        \centering
        \includegraphics[width=\textwidth]{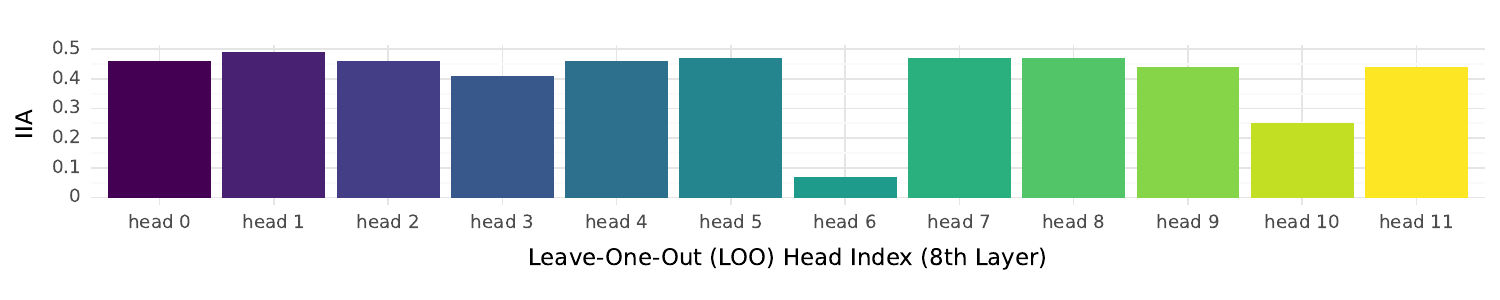}
    \end{minipage}
    \begin{minipage}{1.0\textwidth}
        \centering
        \includegraphics[width=\textwidth]{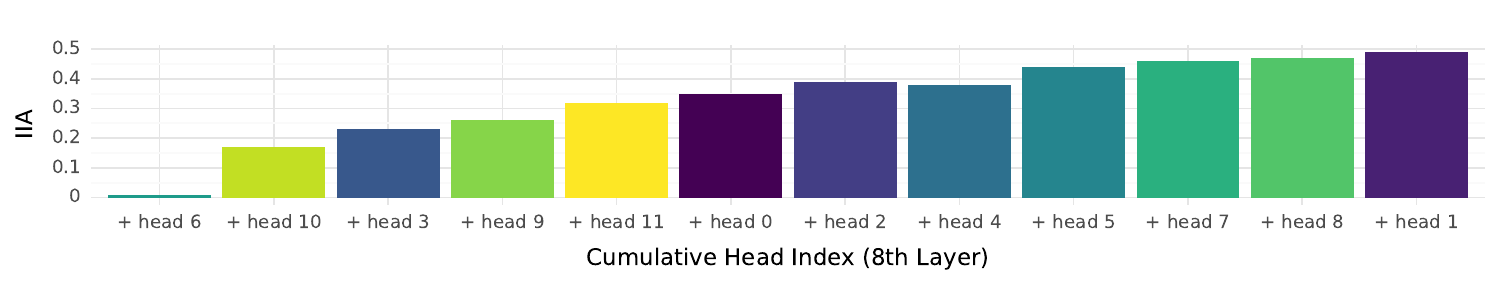}
    \end{minipage}
    \caption{Interchange Intervention Accuracy (IIA) when aligning the name position variable with head representations. The \textbf{top panel} shows IIA when aligning with a concatenated representation of all heads in the 8th layer by leaving one head out at a time. The \textbf{bottom panel} shows IIA when aligning with cumulated head representations by starting from the head resulting in the largest drop in the top panel and concatenating with one additional head at a time based on the sorted order of IIA drops from the top panel.}
    \label{Fig:head_analyses}
\end{figure*}

\begin{figure*}[t]
    \centering
    \begin{minipage}{0.95\textwidth}
        \centering
        \includegraphics[width=\textwidth]{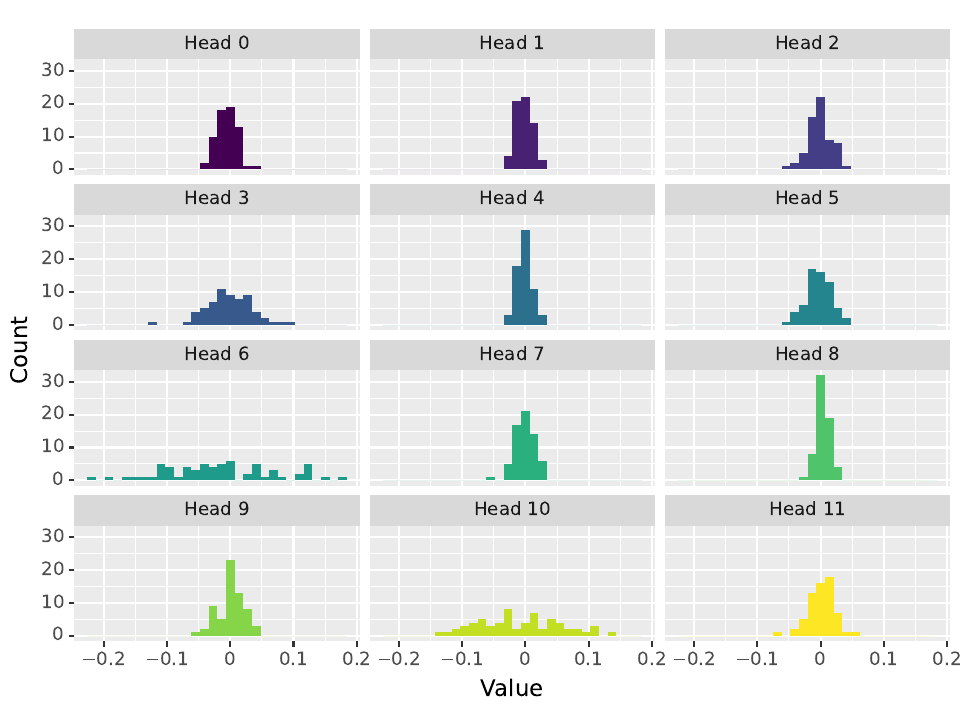}
    \end{minipage}
    \caption{Distributions of learned DAS weights (a single dimension DAS) when aligning with the name position information at attention value output stream of the 8th layer. The number of non-zero entries maps well to the head importance discovered through our ablation studies in \Secref{Sec:head_distribute_reprs} as well as findings from previous works~\citep{wang2022interpretability}.}
    \label{Fig:DAS_weight}
\end{figure*}

\subsection{Name Position Information Is Distributed Across Attention Heads}\label{Sec:head_distribute_reprs}
\citet{wang2022interpretability} show that there are certain heads (e.g., heads 6 and 10) in the 8th layer that are responsible for deriving the name position information. To verify whether a single head can carry the name position information, we train a single DAS direction on the head's representation individually at the 8th layer. Surprisingly, when aligning with each head, IIA flattens out at 0\%, which suggests the name position information may be distributed across multiple heads instead. 

Instead of aligning with individual heads, we perform Leave-One-Out (LOO) alignment: we align with concatenated head representations by leaving one head out at a time. As shown in the top panel of \Figref{Fig:head_analyses}, leaving heads 6 and 10 out results in the largest drops in IIA, which aligns with previous findings by \citet{wang2022interpretability} claiming these two heads are responsible for carrying out the name position information. Nevertheless, we find that the name position information is fairly distributed across heads, or the information emerges only when multiple heads are considered together and heads need to act together to reach good IIA. 

\begin{figure*}[t]
    \centering
    \begin{minipage}{1.0\textwidth}
        \centering
        \includegraphics[width=\textwidth]{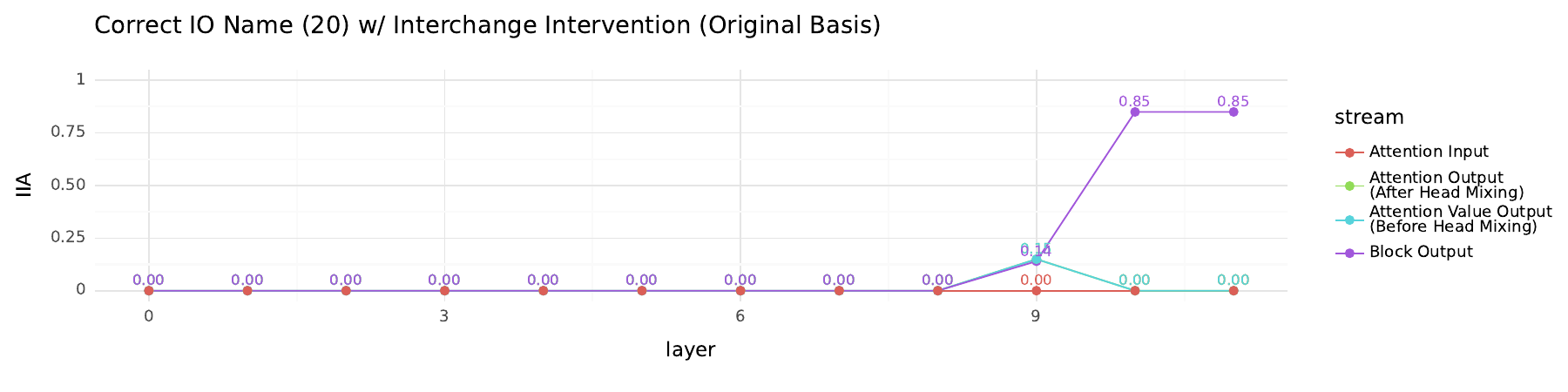}
    \end{minipage}
    \begin{minipage}{1.0\textwidth}
        \centering
        \includegraphics[width=\textwidth]{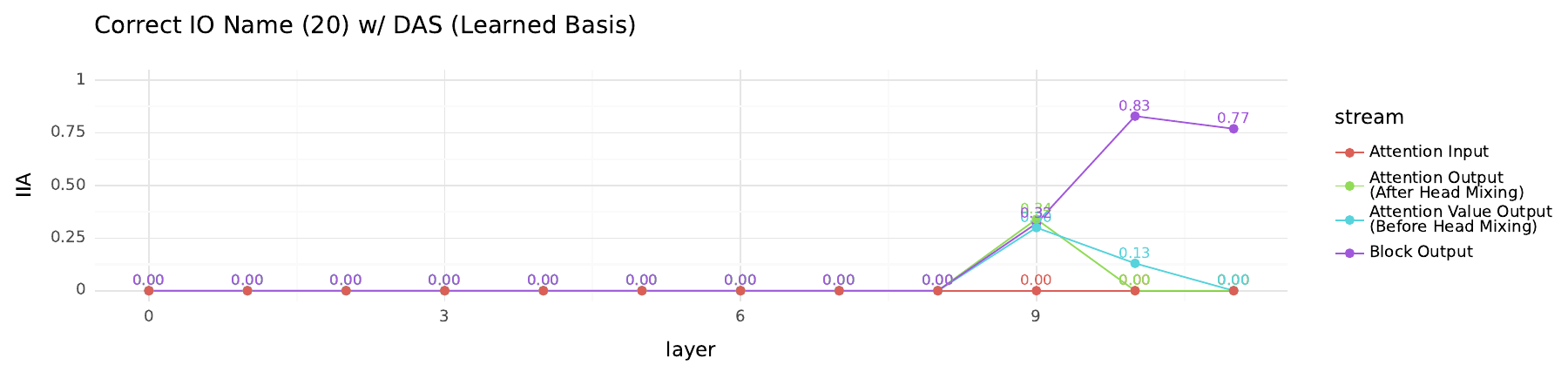}
    \end{minipage}
    \caption{Interchange Intervention Accuracy (IIA) when aligning the correct IO name variable with interchange intervention (i.e., no training) and DAS at different intervention locations above the last token.}
    \label{Fig:IO_name_last_token_streams}
\end{figure*}

\begin{figure*}[h]
    \centering
    \begin{minipage}{1.0\textwidth}
        \centering
        \includegraphics[width=\textwidth]{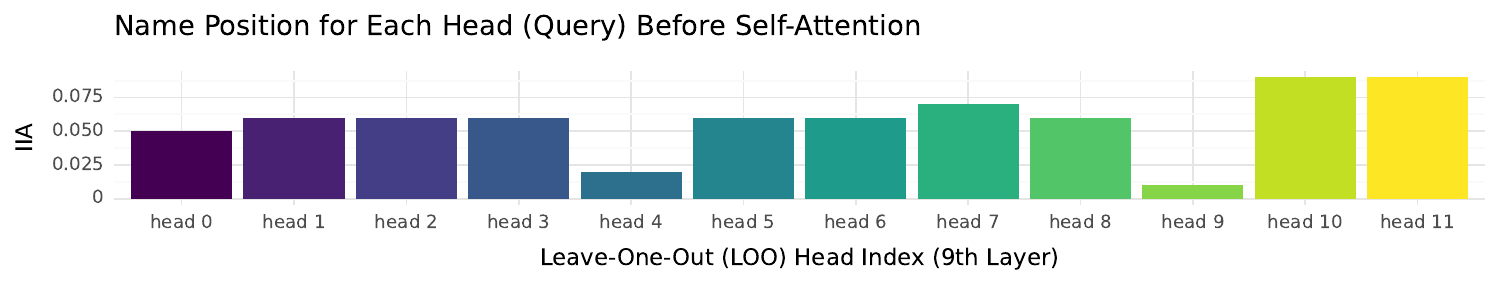}
    \end{minipage}
    \begin{minipage}{1.0\textwidth}
        \centering
        \includegraphics[width=\textwidth]{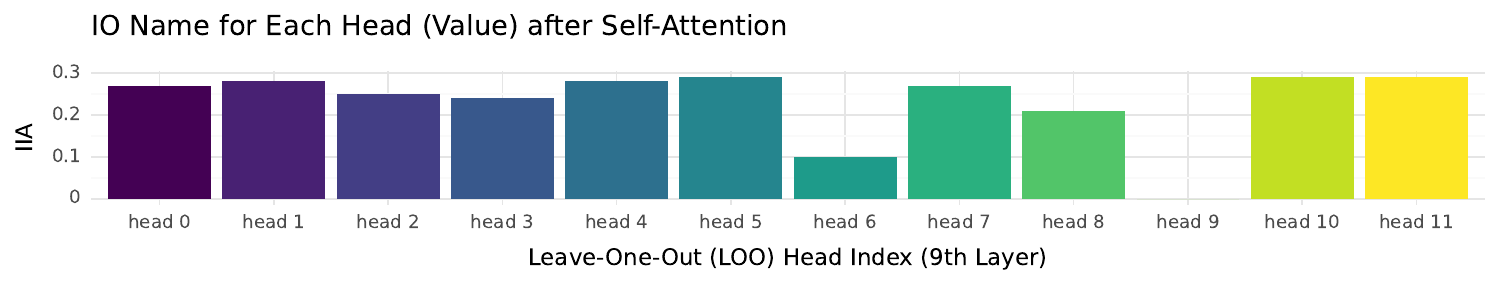}
    \end{minipage}
    \begin{minipage}{1.0\textwidth}
        \centering
        \includegraphics[width=\textwidth]{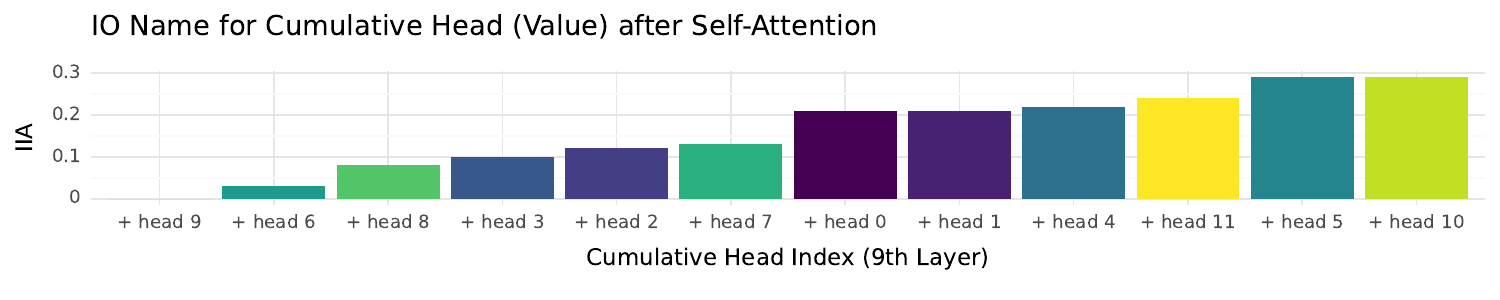}
    \end{minipage}
    \caption{Interchange Intervention Accuracy (IIA) when aligning the name position variable or the correct IO name with head representations at the 9th layer.}
    \label{Fig:head_analyses_for_io_name}
\end{figure*}

To verify this, we then rank the IIA drops across heads from the highest (i.e., leaving this head out results in the biggest loss in IIA) to the lowest based on LOO experiments. We re-align with cumulated head representations by starting from the lowest head itself and concatenating with one additional head at a time based on the sorted order. Our results show that head 6 alone reaches 0\% IIA, and it reaches less than 20\% when considered together with head 10 as shown in the bottom panel of \Figref{Fig:head_analyses}. Our findings suggest that all heads are crucial to carrying out all the name position information. Figure \ref{Fig:DAS_weight} shows the breakdown of the learned DAS weights across heads in the attention value output stream ($v_{\text{attn\_value\_output}}$), without head mixing. Our results indicate that the importance of each head correlates well with the number of non-zero entries in the learned DAS weight matrix. Additional analyses at 7th layer can be found in \Appref{App:additional_head_analyses}. 

\subsection{Name Position Information Is Processed in a Distributed Way}\label{Sec:name_mover_heads}
We further check where the correct IO token emerges above the last token. To find the IO token representation, we use vanilla interchange interventions~\citep{Geiger:2021} as well as DAS. As shown in \Figref{Fig:IO_name_last_token_streams}, the correct IO token seems to emerge starting from the 9th layer. Comparing results from DAS with vanilla interchange intervention, we find that DAS finds new alignments in the attention value output stream with an IIA of 13\% where the vanilla interchange intervention flats out with a 0\% IIA. 

One hypothesis is that self-attention heads at the 9th and 10th layers process the name position information passed in and fetch the correct IO name by adding it to the main residual stream. Indeed, \citet{wang2022interpretability} shows head 9 at the 9th layer is identified as the strongest ``name mover head'' that copies the IO name from the sequence into the last token. To verify this, we align the name position information in the query output stream ($v_{\text{attn\_input}}$) at each head as well as the IO name in the attention value output at each head. Instead of aligning each head individually, we first perform the LOO experiment as in \Secref{Sec:head_distribute_reprs}. \Figref{Fig:head_analyses_for_io_name} shows our results. We find that head 9 indeed absorbs the name position information the most compared to other heads (i.e., removing head 9 results in the largest IIA drops when aligning the name position before self-attention) and contributes to generating the IO name in the attention value output stream (i.e., removing head 9 results in the largest IIA drops when aligning the IO name after self-attention). However, our results also suggest that head 9 alone cannot produce the correct IO name result in 0\% IIA when aligning the IO name by itself. We need up to 11 heads to reach approximately the ceiling IIA of aligning the IO name from the attention value output stream. Our findings suggest that attention heads process information in a very distributed way. Interestingly, the correct IO name alignments disappear after head mixing at 10th layer, suggesting another causal variable may emerge to add the correct IO name information into the main residual stream. \Appref{App:additional_head_analyses} includes additional analyses at 10th layer which shows similar results.

\section{Conclusions}
We disagree with \citet{aleks2023subspace}'s characterization of their examples and results, but we find their contribution to be extremely valuable, and we applaud their effort to probe methods like DAS very deeply. Overall, their work has helped us to more deeply understand DAS as well as how neural networks work in general, and thus we feel the discussion has furthered the goals of explainable AI in general. 

\section*{Acknowledgments}
We thank Aleksandar Makelov, George Lange, and Neel Nanda for providing feedback on an earlier version of this paper.

\bibliography{iclr2024_conference}
\bibliographystyle{iclr2024_conference}

\appendix
\section*{Appendix}

\section{Multiple Causal Abstractions Exist in the Toy Example}\label{App:multi-abstractions}
Besides revisiting the ``illusion'' in the toy example, we also find there is an interesting connection between the disconnected or dormant direction and the existence of multiple causal abstractions.

\subsection{Set-up}
In the original basis, the neural model is implemented as,
\begin{equation}
f(x) = 0\times(1\times x)_{\text{H1}} + 2\times(0\times x)_{\text{H2}} + 1\times(1\times x)_{\text{H3}} = x
\end{equation}
Now, let’s say we have a rotation matrix $\mathbf{R}$ (change-of-basis matrix) as,
\[
\mathbf{R}
= 
\begin{bmatrix}
    \frac{1}{\sqrt{2}} & \frac{1}{\sqrt{2}} & 0\\
    \frac{-1}{\sqrt{2}} & \frac{1}{\sqrt{2}} & 0\\
    0 & 0 & 1\\
\end{bmatrix}
\]
Importantly, we use the subspace identified as an ``illusion'' by \citet{aleks2023subspace} as the first direction in this rotation matrix. We can then rewrite $f(x) = g(x) = W_2^\intercal R^\intercal(RxW_1)$ given $R^{-1} = R^\intercal$ for orthonormal matrix. Thus, $g(x)$ can be expressed as,
\begin{equation}
g(x) = \frac{1}{\sqrt{2}}\times(\sqrt{2}\times x)_{\text{H1}} - \frac{1}{\sqrt{2}}\times(\sqrt{2}\times x)_{\text{H2}} + 1\times(1\times x)_{\text{H3}} = x
\end{equation}

\subsection{Multiple Causal Abstractions}
$f(x)$ and $g(x)$ are behaviorally identical yet structurally different due to the basis change. Likewise, $g(x)$ gives us additional intervention signals when intervening on H1 in the rotated basis. In other words, if we want to align the representation of $x$ (i.e., the input identity) with some representations among \{H1, H2, H3\}, it could be H3 in the original basis by $f(x)$, or H1 in the rotated basis by $g(x)$. This resembles the case where there are multiple causal abstractions when aligning causal variables. We discussed this extensively in \citealt{geiger2023finding} and \citealt{wu-etal-2023-Boundless-DAS}. In essence, the causal abstraction framework cannot determine structural equivalence between objects. Instead, it says that, under a set of known interventions drawn from the data distribution, we cannot distinguish two objects (e.g., a Python program and a neural network) behaviorally under any combinations of these interventions. One could try to align components between two objects more exhaustively to distinguish multiple causal abstractions. For instance, we could further align all hidden representations (e.g., intervention on H2 in $f(x)$ and $g(x)$ will give different counterfactual behavior) in the toy example. However, for sufficiently complex models, there are still likely to be multiple distinct fully correct abstractions. Additionally, there is a correspondence with the complexity of the high-level model. This means that more accurate abstractions are likely when the low-level model is complex, and the high-level model is simple.

\section{Experimental Setup}\label{App:experiment-setup}
We use the GPT-2 model Small checkpoint from HuggingFace.\footnote{GPT-2 can be accessed publicly at \url{https://huggingface.co/gpt2}.} We use the dataset generation scripts from the publicly released code of \citet{aleks2023subspace}.\footnote{The code is publicly released at \url{https://github.com/amakelov/activation-patching-illusion}.} To train a single DAS direction, we sample 200 pairs of base and source example pairs. We use the \texttt{Adam} optimizer \citep{kingma2014adam} with an initial learning rate of $0.01$ for an epoch number of 10 with a batch size of~20. As in \citet{aleks2023subspace}, examples in the dataset have three different templates, yet the base and the source examples in each pair share the same template. Two templates are used during training while an unseen template is used during evaluation. Each input example has the same sequence length of~18. In case the predicting name is tokenized into multiple tokens, the accuracy is calculated using only the first token. Training of a single DAS direction takes less than a minute using a single 12G Nvidia GPU. 

To find the IO name representations, we sample 200 pairs of base and source randomly and have the correct counterfactual label to be the source IO name. Since DAS is likely to overfit to seen IO names, we have 20 distinct IO names as the possible output label during training and testing. We train DAS with a dimension size of 20. We keep other training settings the same as our name position experiments above. For our cumulative head experiments, we sort heads based on their original indices for stability.\footnote{Ideally, the order should not matter. In practice, given the scale of the dataset and the magnitude of IIAs, we observe variance in results as the order changes.} For Boundless DAS, we initialize the dimensionality to occupy half of the total hidden representation size (e.g., a size of 384 when aligned with the block output with a size of 768). The learning rate for boundary learning is set at $0.05$. Furthermore, we apply a weighted loss for Boundless DAS, assigning a weight of 1.0 to the DAS loss and 2.0 to the boundary loss. We set the initial temperature for boundary learning to be 50.0, and the end temperature to be 0.1 with a linear temperature annealing throughout the training.

\begin{figure*}[h]
    \centering
    \begin{minipage}{1.0\textwidth}
        \centering
        \includegraphics[width=\textwidth]{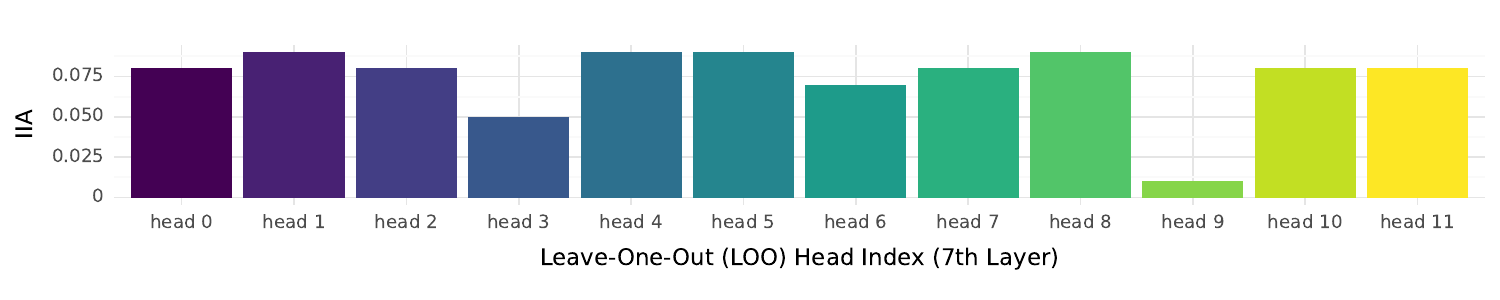}
    \end{minipage}
    \begin{minipage}{1.0\textwidth}
        \centering
        \includegraphics[width=\textwidth]{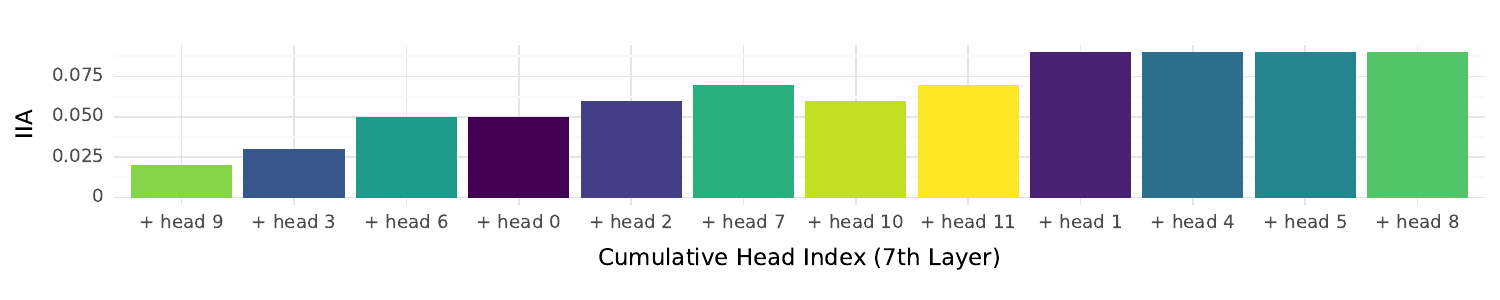}
    \end{minipage}
    \caption{Interchange Intervention Accuracy (IIA) when aligning the name position variable with head representations. The \textbf{top panel} shows IIA when aligning with a concatenated representation of all heads in the 7th layer by leaving one head out at a time. The \textbf{bottom panel} shows IIA when aligning with cumulated head representations by starting from the head resulting in the largest drop in the top panel and concatenating with one additional head at a time based on the sorted order of IIA drops from the top panel.}
    \label{Fig:additional_head_analyses}
\end{figure*}

\begin{figure*}[h]
    \centering
    \begin{minipage}{1.0\textwidth}
        \centering
        \includegraphics[width=\textwidth]{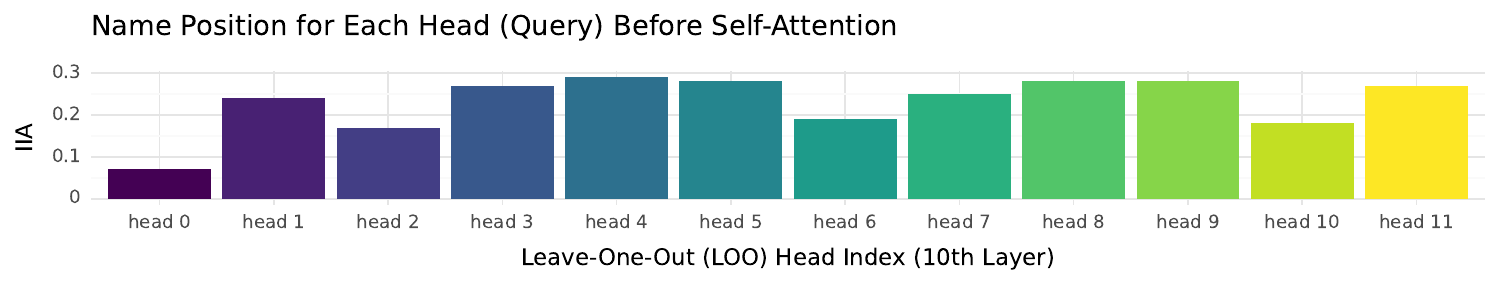}
    \end{minipage}
    \begin{minipage}{1.0\textwidth}
        \centering
        \includegraphics[width=\textwidth]{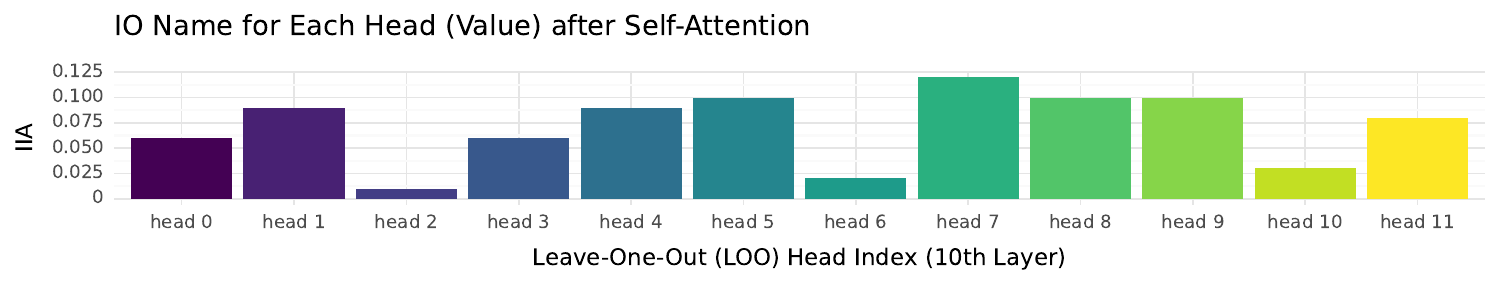}
    \end{minipage}
    \begin{minipage}{1.0\textwidth}
        \centering
        \includegraphics[width=\textwidth]{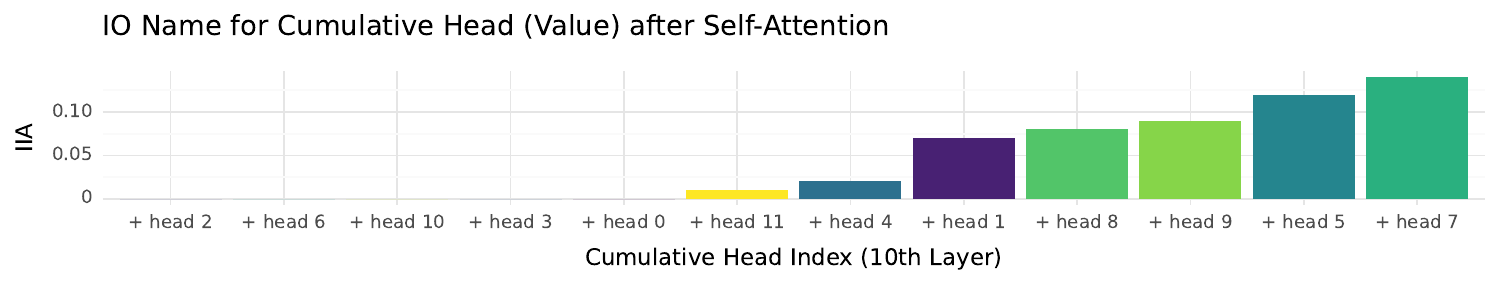}
    \end{minipage}
    \caption{Interchange Intervention Accuracy (IIA) when aligning the name position variable or the correct IO name with head representations at the 10th layer.}
    \label{Fig:additional_head_analyses_for_io_name}
\end{figure*}

\section{Additional Head Alignment Results}\label{App:additional_head_analyses}
We conduct the same set of analyses outlined in \Secref{Sec:head_distribute_reprs} at the 7th layer as well since our results in \Figref{Fig:last_token_streams} show that attention value representations in the 7th layer also carry some name position information. As shown in \Figref{Fig:additional_head_analyses}, there are particular heads (e.g., heads 9 and 3) that result in larger IIA drops compared to others when removed. Similarly, the name position information is fairly distributed across heads. Notably, removing heads in \{4, 5, 8\} results in almost no drop in IIA suggesting information is less distributed compared to the 8th layer. \Figref{Fig:additional_head_analyses_for_io_name} shows additional results when aligning the name position variable or the correct IO name with head representations at the 10th layer. Figure \ref{Fig:DAS_IO_name_weight} shows the distributions of the learned DAS weights for each head when aligning the correct IO name with the attention value output of the 9th layer. Consistent with our findings in Section \ref{Sec:head_distribute_reprs}, the importance of each head corresponds well with the number of non-zero entries in the learned weight matrix.

\section{Additional Boundless DAS results}\label{App:additional_boundless_das_analyses}

\begin{figure*}[t]
    \centering
    \begin{minipage}{1.0\textwidth}
        \centering
        \includegraphics[width=\textwidth]{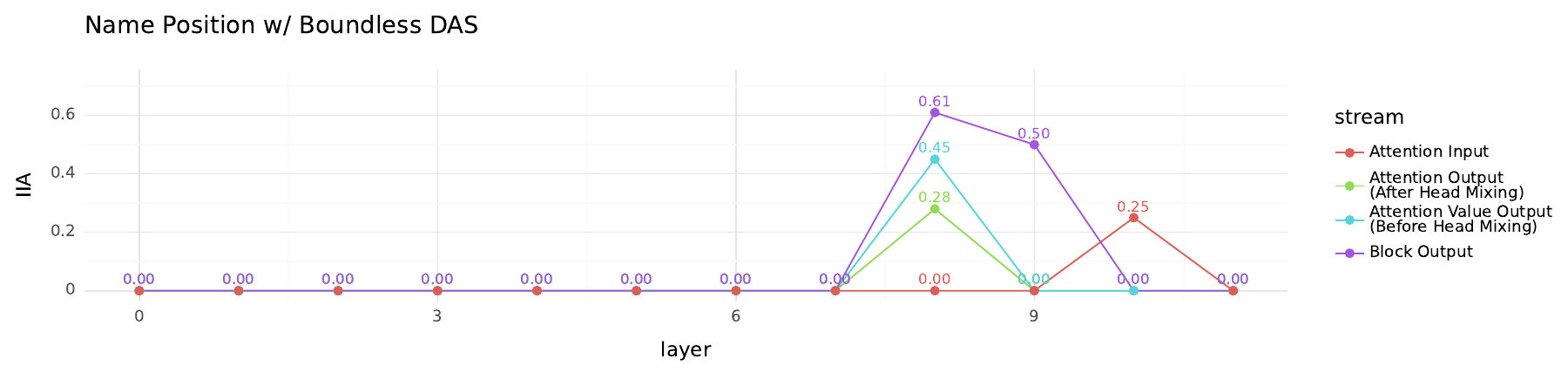}
    \end{minipage}
    \begin{minipage}{1.0\textwidth}
        \centering
        \includegraphics[width=\textwidth]{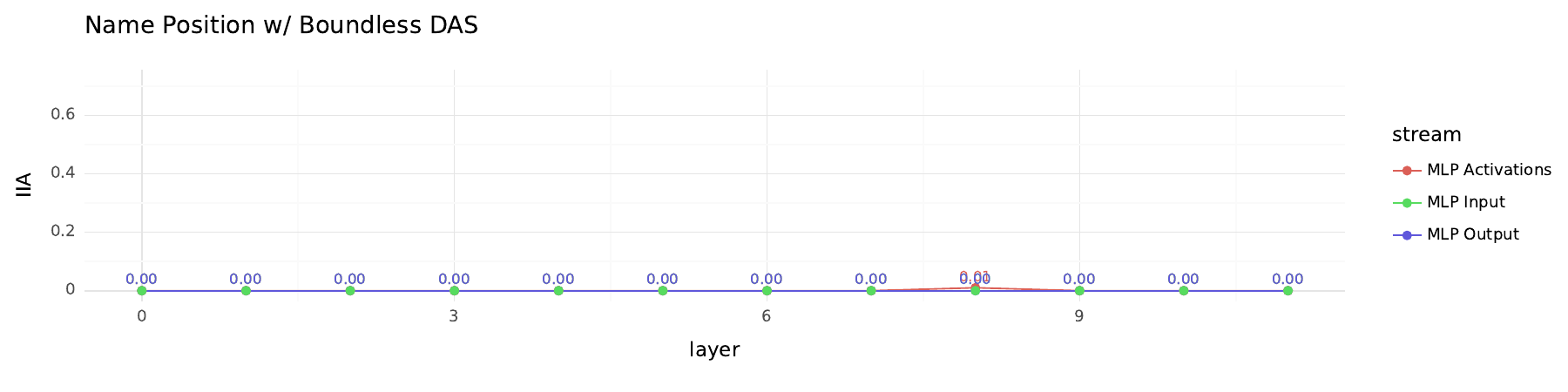}
    \end{minipage}
    \caption{Interchange Intervention Accuracy (IIA) when aligning the name position variable with Boundless DAS and at different intervention locations above the last token.}
    \label{Fig:BDAS_name_position_last_token_streams}
\end{figure*}

\begin{figure*}[t]
    \centering
    \begin{minipage}{1.0\textwidth}
        \centering
        \includegraphics[width=\textwidth]{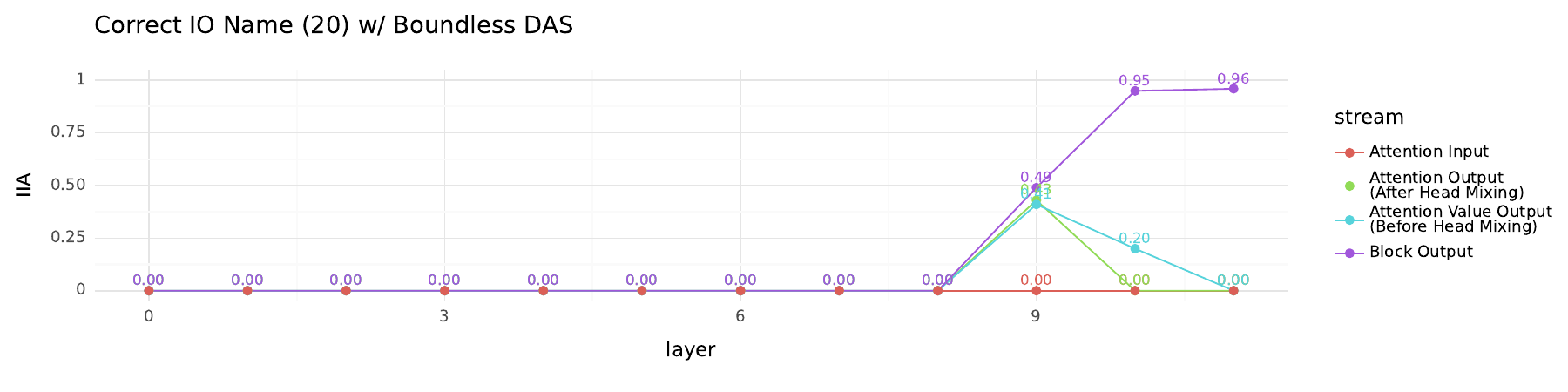}
    \end{minipage}
    \begin{minipage}{1.0\textwidth}
        \centering
        \includegraphics[width=\textwidth]{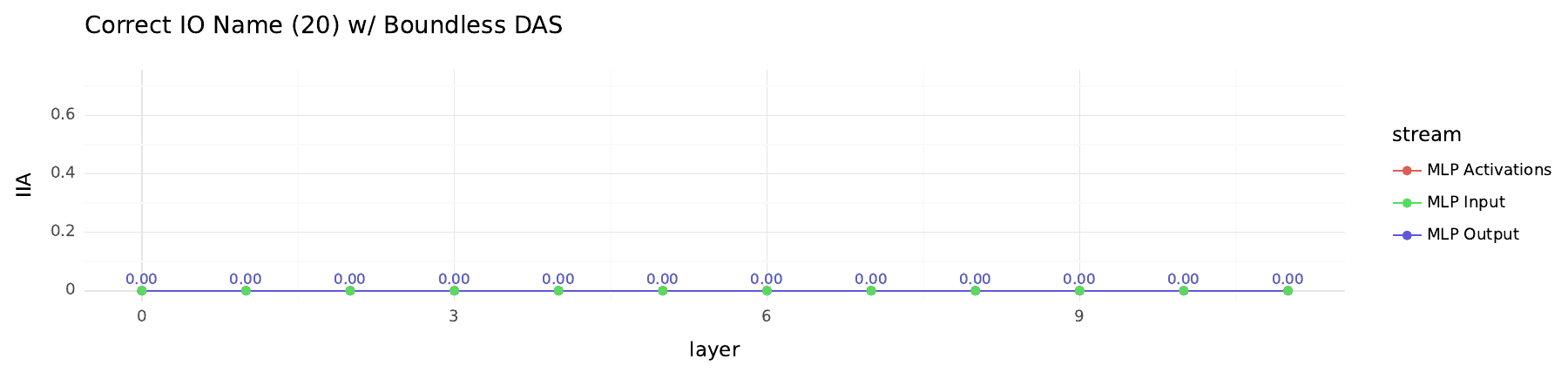}
    \end{minipage}
    \caption{Interchange Intervention Accuracy (IIA) when aligning the correct IO name variable with Boundless DAS and at different intervention locations above the last token.}
    \label{Fig:BDAS_IO_name_last_token_streams}
\end{figure*}

\begin{figure*}[h]
    \centering
    \begin{minipage}{1.0\textwidth}
        \centering
        \includegraphics[width=\textwidth]{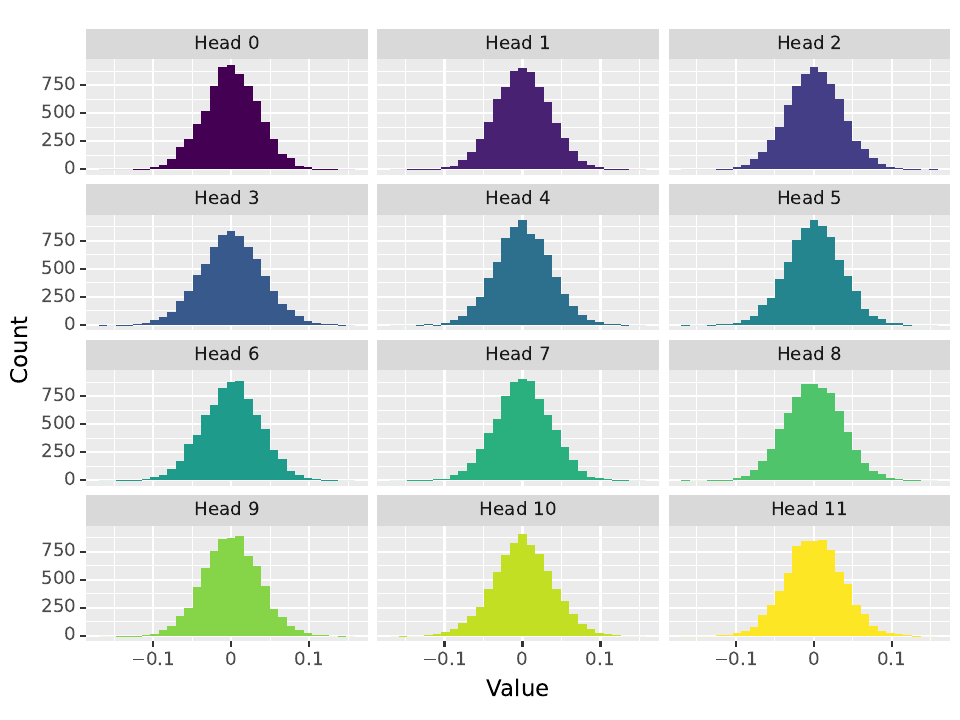}
    \end{minipage}
    \caption{Distributions of learned Boundless DAS weights (learned boundary occupied 14.88\% of the original dimension size) when aligning with the name position information at attention value output stream of the 8th layer.}
    \label{Fig:Boundless_DAS_weight}
\end{figure*}

\begin{figure*}[h]
    \centering
    \begin{minipage}{1.0\textwidth}
        \centering
        \includegraphics[width=\textwidth]{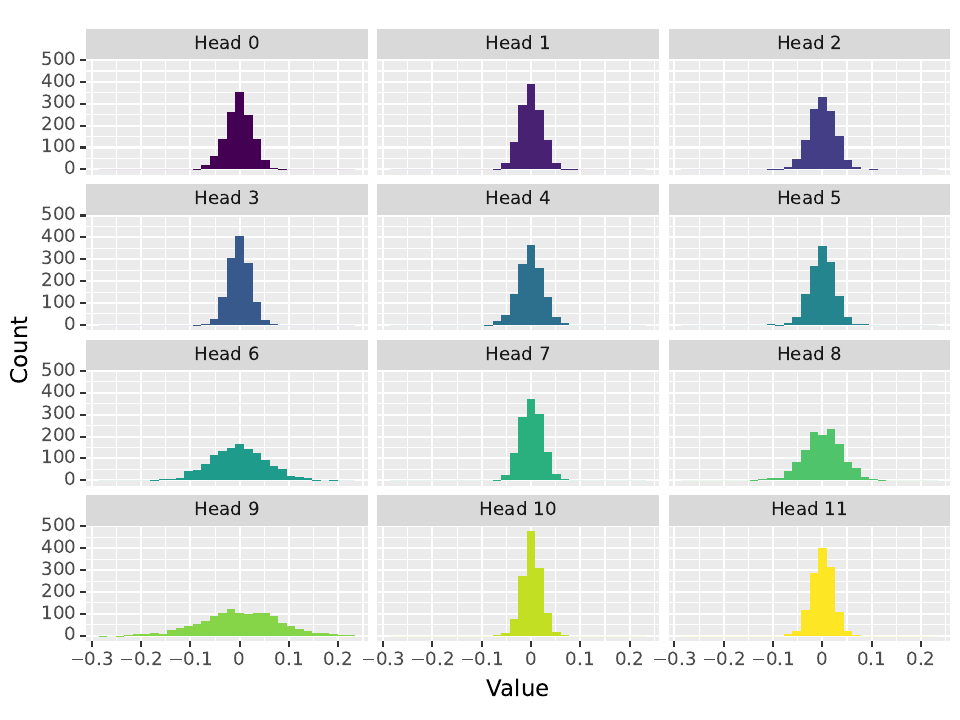}
    \end{minipage}
    \caption{Distributions of learned DAS weights (a fixed dimensionality of 20) when aligning with the correct IO name at attention value output stream of the 9th layer.}
    \label{Fig:DAS_IO_name_weight}
\end{figure*}

\begin{figure*}[h]
    \centering
    \begin{minipage}{1.0\textwidth}
        \centering
        \includegraphics[width=\textwidth]{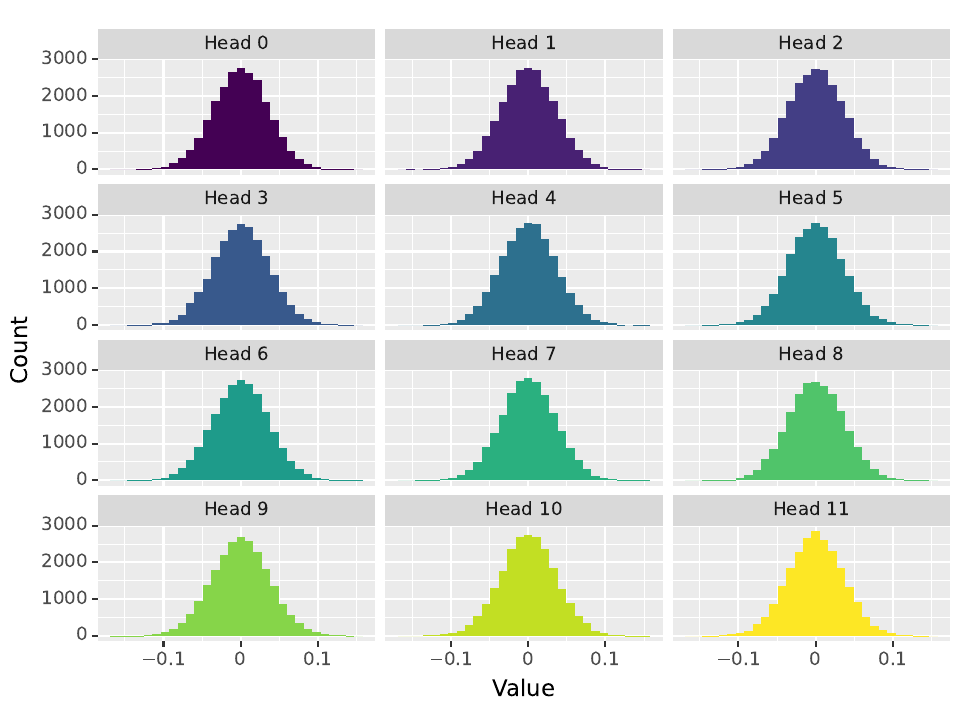}
    \end{minipage}
    \caption{Distributions of learned Boundless DAS weights (learned boundary occupied 46.87\% of the original dimension size) when aligning with the correct IO name at attention value output stream of the 9th layer.}
    \label{Fig:Boundless_DAS_IO_name_weight}
\end{figure*}

In \Figref{Fig:BDAS_name_position_last_token_streams} and \Figref{Fig:BDAS_IO_name_last_token_streams}, additional alignment results are presented using Boundless DAS~\citep{wu-etal-2023-Boundless-DAS}. Boundless DAS achieves comparable outcomes to DAS when aligning the name position variable. However, it demonstrates superior performance in alignment with the correct IO name. \Figref{Fig:Boundless_DAS_weight} shows the distributions of learned DAS weight for each head when aligning the name position information with the attention value output of the 8th layer. Compared with DAS results in \Figref{Fig:DAS_weight}, contrasts between heads are much less salient, suggesting that Boundless DAS learns a much more distributed subspace. \Figref{Fig:Boundless_DAS_IO_name_weight} shows the distributions of learned DAS weight for each head when aligning the correct IO name with the attention value output of the 9th layer, which suggests a similar finding.

\end{document}